\documentclass{llncs} 
\usepackage{hyperref}
\usepackage{url}
\usepackage{cite}
\usepackage[pdftex]{graphicx}
\usepackage[cmex10]{amsmath}
\usepackage{amsfonts}
\usepackage[noend]{algorithmic}
\usepackage{array}
\usepackage{multirow}
\usepackage{graphicx}
\usepackage{caption}
\usepackage{subcaption}
\usepackage[]{algorithm2e}
\usepackage{verbatim}
\usepackage{perpage}
\usepackage{booktabs}
\usepackage{multirow}

\MakePerPage{footnote} 

\urldef{\mailsa}\path|nastaran.mohammadianrad@unitn.it|
\newcommand{\keywords}[1]{\par\addvspace\baselineskip
\noindent\keywordname\enspace\ignorespaces#1}

\begin{document}

\title{Convolutional Neural Network for Stereotypical Motor Movement Detection in Autism}
\author{Nastaran Mohammadian Rad\inst{1,2}\and Andrea Bizzego\inst{1,2}\and Seyed Mostafa Kia\inst{1,2}\and Giuseppe Jurman\inst{2}\and Paola Venuti\inst{1} \and Cesare Furlanello\inst{2}}
\institute{University of Trento, Trento, Italy \and Bruno Kessler Foundation, Trento, Italy}
\maketitle
\begin{abstract}

Autism Spectrum Disorders (ASDs) are associated with specific atypical postural or motor behaviors, of which Stereotypical Motor Movements (SMMs) may severely interfere with learning and social interactions. Wireless inertial sensing technology offers a valid infrastructure for automatic and real-time SMM detection, which would provide support for tuned intervention and possibly early alert on the onset of meltdown events. However, the identification and the quantification of SMM patterns remains complex due to strong inter-subject and intra-subject variability, hard to deal with by handcrafted features. Here we propose to employ the deep learning paradigm in order to learn discriminative features directly from multi-sensor accelerometer signals. Our results with convolutional neural networks provide preliminary evidence that feature learning and transfer learning embedded in deep architectures may lead to accurate and robust SMM detectors in longitudinal scenarios.
\end{abstract}


\vspace{-1em}
\keywords{Autism, SMM, Deep learning, Wireless inertial sensors.}
\vspace{-1em}
\section{Introduction}
\label{sec:introduction}
Autism Spectrum Disorders (ASDs) are defined as a range of developmental disability conditions that effect at some degree the social interaction and communication abilities of patients. Prevalence of ASD is reported to be 1 in 88 individuals~\cite{baio2012prevalence}. ASD is generally characterized by restricted, repetitive, and stereotyped patterns of behavior in patients. Stereotypical Motor Movements (SMMs) in autism (such as body rocking, mouthing, and complex hand movements)~\cite{lagrow1984stereotypic} can significantly restrict the learning and social interactions. Further, SMMs frequently increase in occasion of emotional or sensory overload, which may lead to autistic meltdown events. Alleviating SMMs is thus one primary target of interventions on ASD, which requires accurate tools for recognizing and quantifying SMM patterns. In order to guide behavioral interventions~\cite{goodwin2014moving} and possibly prevent SMM insurgence, it is worthwhile to consider the limitations of traditional methods for measuring SMM~\cite{goodwin2014moving}, e.g., paper-and-pencil rating scales, direct behavioral observation, and video-based coding. Measures by wireless accelerometer sensing technology and machine learning techniques provide an automatic, time efficient, and accurate measure of SMM~\cite{westeyn2005recognizing,min2009optimal,min2010automatic,min2010novel,gonccalves2012automatic,gonccalves2012automatic1,
rodrigues2013stereotyped,albinali2012detecting,albinali2009recognizing,goodwin2011automated,goodwin2014moving,plotz2012automatic}. 
Research by Goodwin et al.~\cite{goodwin2014moving} showed the potential of automatic SMM detection in real-life settings and the limits towards developing robust and adaptive real-time algorithms.

As in many other signal processing applications, SMM detection is commonly based on extracting \emph{ad-hoc} (``handcrafted") features from the accelerometer signals. So far, a wide variety of feature extraction methods have been used in the literature. Generally, two types of features are extracted from the accelerometer signal~\cite{ince2008detection}: i) time domain features; ii) frequency domain features. For time domain features, statistical features such as mean, standard deviation, zero-crossing, energy, and correlation are extracted from overlapping windows of signal. In frequency domain features, the discrete Fourier transform is used to estimate the power of different frequency bands. Recently, the Stockwell transform~\cite{stockwell1996localization} has been proposed for feature extraction from inertial 3-axis accelerometer in order to provide better time-frequency resolution for non-stationary signals~\cite{goodwin2014moving}. Despite its popularity in movement analysis, manual feature extraction suffers from two main limitations~\cite{martinez2013learning}: a) the feature extraction phase is mainly based on generic researchers' domain knowledge rather than encoding movement information. Thus, characteristics of atypical movements may be missed, without coping with intra-subject and inter-subject variation; b) feature extraction is a computationally intensive step in the processing pipeline and such computational cost limits the applicability of atypical movement detection in real-time scenarios.

To overcome such limitations, we propose to use the deep learning paradigm in order to learn discriminating features for SMM pattern recognition. In particular, we introduce a Convolutional Neural Network (CNN) deep learning model~\cite{lecun1995convolutional} to bypass the commonly used feature extraction procedure. The CNN can be employed to transform the multi-channel accelerometer signal into a reduced set of features, and then an SVM classifier is used to classify the new representation of signal into SMM and no-SMM classes. The CNN architecture is shown to be an effective machine learning technique for a wide range of problems such as object recognition~\cite{szegedy2014going,lawrence1997face}, speech processing~\cite{abdel2012applying}, and affect recognition~\cite{martinez2013learning}. We hypothesize that the feature learning and transfer learning capabilities of CNN provide more accurate SMM detectors, as well as a platform for learning more robust representation of inertial signals, thus giving the capability of effectively transforming this learned representation to a new dataset, which is essential in longitudinal studies. To the best of the authors' knowledge, the CNN model has not been applied in SMM detection applications so far.

   

\section{Methods}
\label{sec:methods}
\begin{figure}
         \centering
         \begin{subfigure}[h!]{0.8\textwidth}
                 \includegraphics[width=\textwidth]{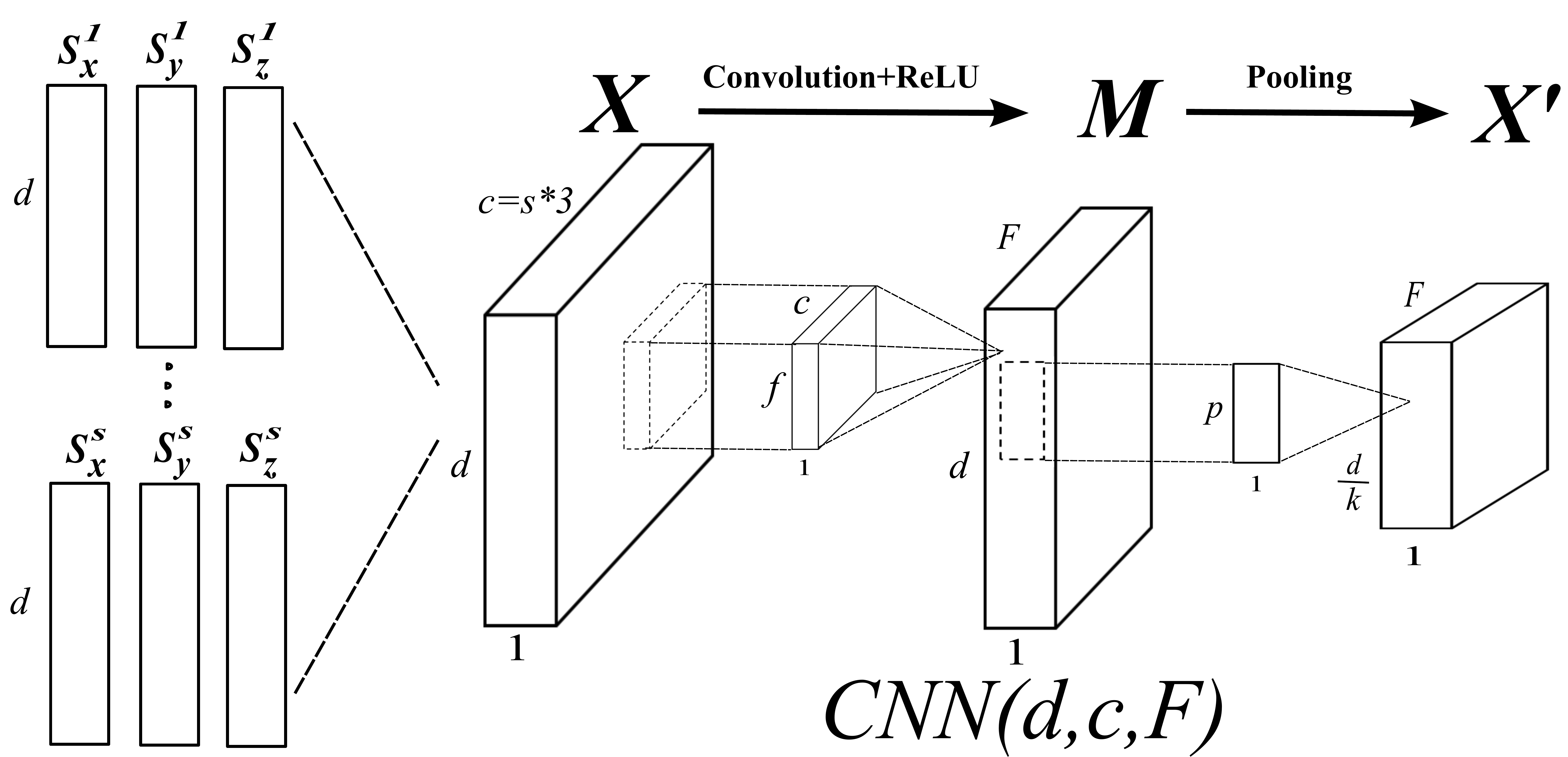}
         \end{subfigure}
         \caption{One layer of CNN, $CNN(d,c,F)$, where $d$ is the length of input signal, $c$ is the number of input channels, and $F$ is the number of filters.} 
         \label{fig:CNNLayer}       
\end{figure}

Let $\textbf{S}_x^i,\textbf{S}_y^i,\textbf{S}_z^i \in \mathbb{R}^{n \times d}$ be $n$ samples of recorded signal by $i$th $\in \{1,2,...,s\}$ accelerometer sensor with $d$ sampling rate at $x$,$y$, and $z$ directions, respectively. Assume $Y^{n \times 1} \in \{-1,1\}$ be the corresponding label vector for the recorded data where $-1$ and $1$ represent \emph{no-SMM} and \emph{SMM} classes, respectively. Then let $\textbf{X} \in \mathbb{R}^{n \times c \times d}$ be a 3-dimensional tensor matrix constructed by concatenating the signal of $s$ accelerometer sensors along each sensor-direction dimension (see Figure~\ref{fig:CNNLayer}), where $c=s*3$ (3 is the number of directions, i.e. $x$,$y$, and $z$). With this simple encoding, we consider each direction of a sensor as an input data channel. 
\vspace{-2em}
\subsection{Feature Learning via Convolutional Neural Network}
\label{subsec:CNN}
Convolutional neural networks (CNNs) benefit from invariant local receptive fields, shared weights, and spatio-temporal sub-sampling features to provide robustness over shift and distortion of the input space~\cite{lecun1995convolutional}. A typical CNN has a hierarchical architecture that alternates convolutional and pooling layers in order to summarize large input spaces with spatio-temporal relations into a lower dimensional feature space. A 1-dimensional convolutional layer $\mathcal{C}^i$ contains a set of $F$ filters, i.e. receptive fields, $\Phi^i = \left \{ \mathcal{F}^i_j \in \mathbb{R}^f \mid j \in \{1,2,\dots,F \} \right \}$ which learn different patterns on a time window of time-series, where $f$ represents the size of each filter. The aim of the training phase is indeed to learn these filters from the input data. Each filter is convolved sequentially with the input signal across channels. Then, the output of the convolution operator is passed through an activation function to compute the feature maps $\textbf{M} \in \mathbb{R}^{n \times F \times d}$. Generally, a rectified linear unit (ReLU) is used as an activation function in a deep architecture, where $ReLU(a) = max\{0,a\}$. To reduce the sensitivity of the output to shifts and distortions, feature maps are fed to an additional layer, called \emph{pooling layer}, which performs a local averaging or sub-sampling. In fact, a pooling layer reduces the resolution of a feature map by factor of $\frac{1}{k}$ where $k$ is the stride size, i.e. the stride between successive pooling regions. Max-pooling and average-pooling are two commonly used pooling functions which compute maximum or average among values in a pooling window, respectively. This aggregation is separately performed inside each feature map and provides $\textbf{X}' \in \mathbb{R}^{n \times F \times \frac{d}{k}}$ as the output of the pooling layer. $\textbf{X}'$ can be used as an input to another convolutional layer $\mathcal{C}^{i+1}$ in a multi-layer architecture. Figure~\ref{fig:CNNLayer} illustrates the basic architecture of one layer of a 1-dimensional convolutional layer. 

\subsection{Network Architecture}
\label{subsec:arcitecture}
\begin{figure}
         \centering
         \begin{subfigure}[h!]{0.8\textwidth}
                 \includegraphics[width=\textwidth]{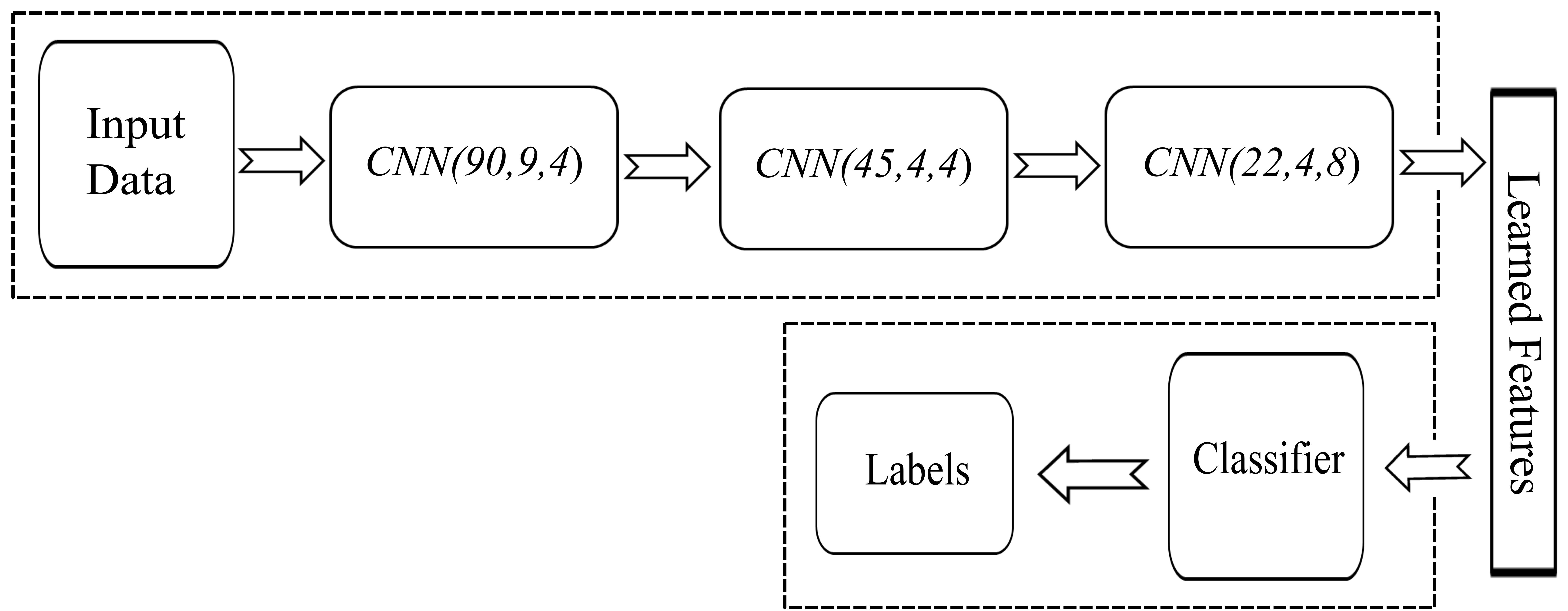}
         \end{subfigure}
         \caption{The CNN architecture for SMM detection.} 
         \label{fig:CNNArchitecture}
\end{figure}

Here we consider a three-layer CNN to transform the time-series of multiple accelerometer sensors into a new feature space. The SMM-detector architecture is shown in Figure~\ref{fig:CNNArchitecture}. Three convolutional layers $\mathcal{C}^1,\mathcal{C}^2,\mathcal{C}^3$ have $4$, $4$, and $8$ filters respectively, each with a length of $9$ time-points ($0.1$ second); i.e., $\Phi^1,\Phi^2= \left \{ \mathcal{F}^{1,2}_j \in \mathbb{R}^9 \mid j \in \{1,\dots,4 \} \right \}$ and $\Phi^3= \left \{ \mathcal{F}^{3}_j \in \mathbb{R}^9 \mid j \in \{1,\dots,8 \} \right \}$. Each convolutional layer is followed by an average-pooling layer. The length of the pooling window and the pooling stride are fixed to $3$ ($p=3$) and $2$, respectively. Pooling stride of $2$ reduces the length of feature maps by factor of $0.5$. The output of the third convolutional layer is connected to a flattening layer to provide the learned feature vector. In the learning phase, a fully-connected layer with $8$ neurons connected to a softmax layer is employed for classification. To configure and train CNN networks, we applied the Deeppy library\footnote{\url{http://andersbll.github.io/deeppy-website/index.html}}, which provides a GPU based infrastructure for computation~\cite{larsen2014cudarray}. 
\vspace{-2em}


\section{Experiments}
\label{sec:experiments}
\subsection{Data and Preprocessing}
\label{subsec:materials}
We used a dataset of accelerometer signals, collected from 6 subjects with autism in a longitudinal study~\cite{goodwin2014moving}\footnote{The dataset and full description of data is publicly available at \url{https://bitbucket.org/mhealthresearchgroup/stereotypypublicdataset-sourcecodes/downloads}.}. The data were collected in laboratory and classroom environments; the subjects wore three 3-axis wireless accelerometer sensors and engaged in SMMs (body rocking, hand flapping, or simultaneous body rocking and hand flapping) and non-SMM behaviors. The sensors were worn on the left wrist and right wrist using wristbands, and on the torso using a thin strip of comfortable fabric tied around the chest. To annotate the data, subject activities were recorded with a video camera and analyzed by an expert. The first data collection (here called \emph{Study1}), were recorded by MITes sensors at $60Hz$ sampling frequency~\cite{albinali2009recognizing}. The second dataset (\emph{Study2}) was collected on the same subjects three years later by Wockets sensors with sampling frequency of $90Hz$. To equalize sampling frequencies between two datasets, Study1 data are resampled to  $90Hz$ with a linear interpolation. To remove DC component from signal, a $0.1Hz$ cut-off high pass filter is applied. Then, similar to~\cite{goodwin2014moving}, the signal is segmented to 1-second long (i.e. 90 time-points) using a sliding window. The sliding window is moved along the time dimension with 10 time-steps resulting in $0.87$ overlap between consecutive windows. Considering the data are collected using 3 sensors, $\textbf{X}$ is a $n \times 9 \times 90$ matrix. Due to the skewness of classes, same as~\cite{goodwin2014moving}, the training data are balanced to the number of samples in the minority class. After constructing the input matrix $\textbf{X}$, Zero Component Analysis (ZCA) is used to normalize the input data. 

\subsection{Experimental Setup}
\label{subsec:setup}
To investigate the effect of feature learning and transfer learning via CNN in SMM detection, we conducted four experiments. In all experiments, the leave-one-subject-out scheme is used for model evaluation. For the sake of fair comparison with~\cite{goodwin2014moving}, Support Vector Machine (SVM) is used for classifying the learned features to target classes.

\vspace{-1.5em}
\subsubsection{Experiment 1:}
The aim of this experiment is to evaluate a baseline for the effect of feature extraction and feature learning on the classification performance. Therefore, without any feature extraction, samples of raw data are used as the input to the SVM classifier for SMM detection. In this case, all data channels of each sample in $\textbf{X}$ are collapsed into a  vector and a $n \times 810$ ($810=9 \times 90$) input matrix is constructed. We will refer to this experiment as \emph{``Raw"}. 
\vspace{-1.5em}
\subsubsection{Experiment 2:}
In this setting, we replicated the third experiment in~\cite{goodwin2014moving} using exactly the same implementation provided by the authors. All extracted features mentioned in~\cite{goodwin2014moving} including time, frequency, and Stockwell transform features are used for the classification. We will refer to this experiment as \emph{``Goodwin et al."}.
\vspace{-1.5em}
\subsubsection{Experiment 3:}
The main aim of this experiment is to investigate the superiority of learning robust features over handcrafted features in the across-subject classification setting. To this end, the proposed CNN architecture is used to learn a middle representation of the accelerometer signals, i.e. to learn the features. In the training phase, one layer of $8$ hidden neurons followed by two softmax neurons (since it is a binary classification problem) are attached densely to the last layer of CNN. All parameters of CNN, i.e. weights and biases, are initialized by drawing small random numbers from normal distribution. The stochastic gradient descent with momentum (fixed to 0.9) is used for training the CNN. Then an SVM classifier is used to classify the new learned feature space to the target labels. All these steps performed on the training set to ensure unbiased error estimation. Due to the random initialization of weights and employing stochastic gradient descent algorithm for optimization, results can be different from one run to another. Therefore, we repeated the whole procedure of learning and classification 15 times, and the errorbars are reported. This experiment is performed separately on Study1 and Study2 data and is referred as \emph{``CNN"}.  
\vspace{-1.5em}
\subsubsection{Experiment 4:} 
In this experiment, we investigate the possibility of transferring learned knowledge from one dataset to another. To this end, we firstly trained the CNN on one dataset, e.g., Study1, and then we used the learned parameters, i.e. filters and weights, for initializing the parameters of CNN in another dataset, e.g., Study2. In fact, we tried to transfer the learned representation from one study to another in a longitudinal study. We refer to this experiment as \emph{``Transferred-CNN"}.

\vspace{-2em}
\begin{table}[]
\centering
\tiny
\caption{F1-score of four experiments on 6 subjects of Study1 and Study2 datasets.}
\label{tab:results}
\begin{tabular}{@{}ccccccccc@{}}
\toprule
\textbf{Study} & \textbf{Experiments} & \textbf{Subj1} & \textbf{Subj2} & \textbf{Subj3} & \textbf{Subj4} & \textbf{Subj5} & \textbf{Subj6} & \textbf{Mean} \\ \midrule
\multicolumn{1}{c|}{\multirow{4}{*}{\textbf{Study1}}} & \textbf{Raw} & 0.43 & 0.26 & 0.18 & 0.44 & 0.56 & 0.56 & 0.41 \\
\multicolumn{1}{c|}{} & \textbf{Goodwin et al.} & 0.73 & 0.36 & 0.5 & 0.73 & 0.44 & 0.46 & 0.54 \\
\multicolumn{1}{c|}{} & \textbf{CNN} & \boldmath$0.74\pm0.02$ & \boldmath$0.75\pm0.02$ & $0.64\pm0.04$ & $0.92\pm0.01$ & $0.51\pm0.04$ & $0.9\pm0.01$ & 0.74 \\
\multicolumn{1}{c|}{} & \textbf{Transferred-CNN} & $0.7\pm0.03$ & $0.74\pm0.01$ & \boldmath$0.69\pm0.03$ & \boldmath$0.92\pm0.006$ & \boldmath$0.68\pm0.06$ & \boldmath$0.93\pm0.01$ & \textbf{0.78} \\ \midrule
\multicolumn{1}{c|}{\multirow{4}{*}{\textbf{Study2}}} & \textbf{Raw} & 0.45 & 0.21 & 0.003 & 0.3 & 0.35 & 0.52 & 0.3 \\
\multicolumn{1}{c|}{} & \textbf{Goodwin et al.} & 0.43 & \textbf{0.26} & \textbf{0.03} & \textbf{0.86} & \textbf{0.72} & 0.07 & 0.4 \\
\multicolumn{1}{c|}{} & \textbf{CNN} & $0.61\pm0.11$ & $0.2\pm0.04$ & $0.02\pm0.007$ & $0.72\pm0.03$ & $0.21\pm0.09$ & $0.36\pm0.13$ & 0.35 \\
\multicolumn{1}{c|}{} & \textbf{Transferred-CNN} & \boldmath$0.67\pm0.08$ & $0.22\pm0.03$ & $0.02\mp0.006$ & $0.75\pm0.03$ & $0.68\pm0.08$ & \boldmath$0.84\pm0.07$ & \textbf{0.53} \\ \bottomrule
\end{tabular}
\end{table}

\vspace{-2em}
\begin{figure}
         \centering
         \begin{subfigure}[b]{1\textwidth}
                 \includegraphics[width=\textwidth]{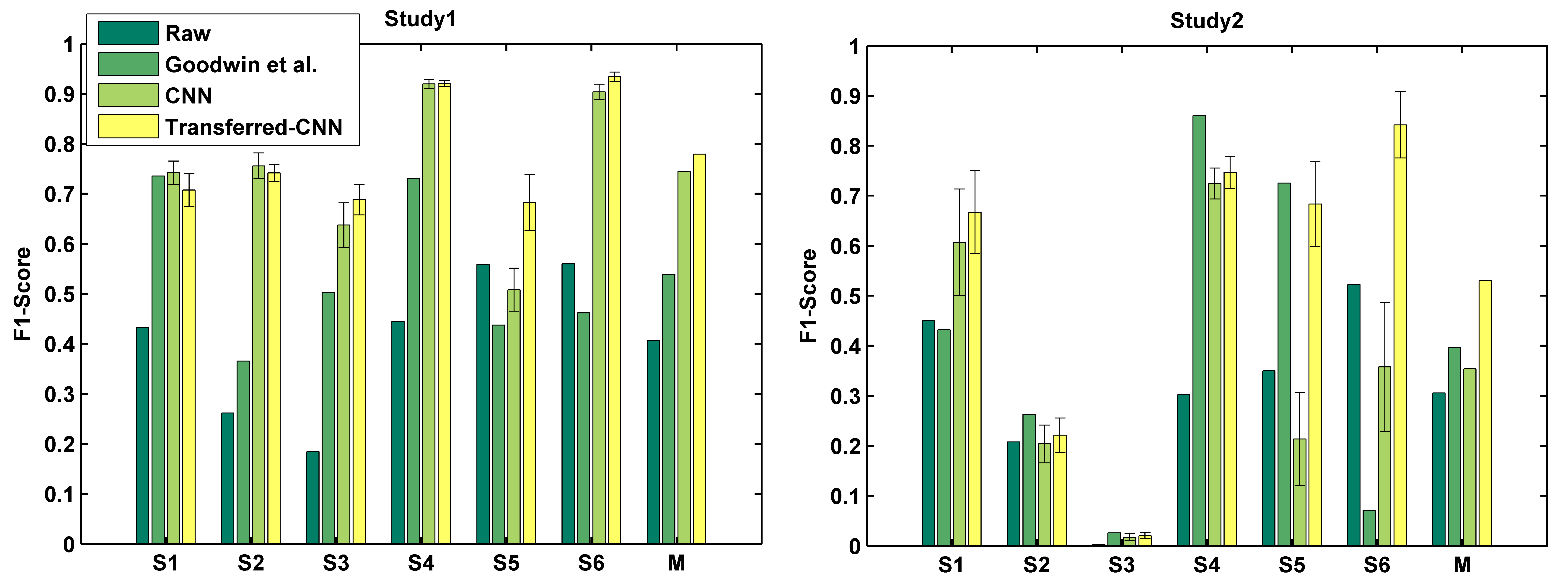}
         \end{subfigure}
         \caption{Comparison between results of four experiments.} 
         \label{fig:results}
\end{figure}
\vspace{-2em}

\subsection{Results and Discussions}
\begin{figure}
         \centering
         \begin{subfigure}[b]{1\textwidth}
                 \includegraphics[width=\textwidth]{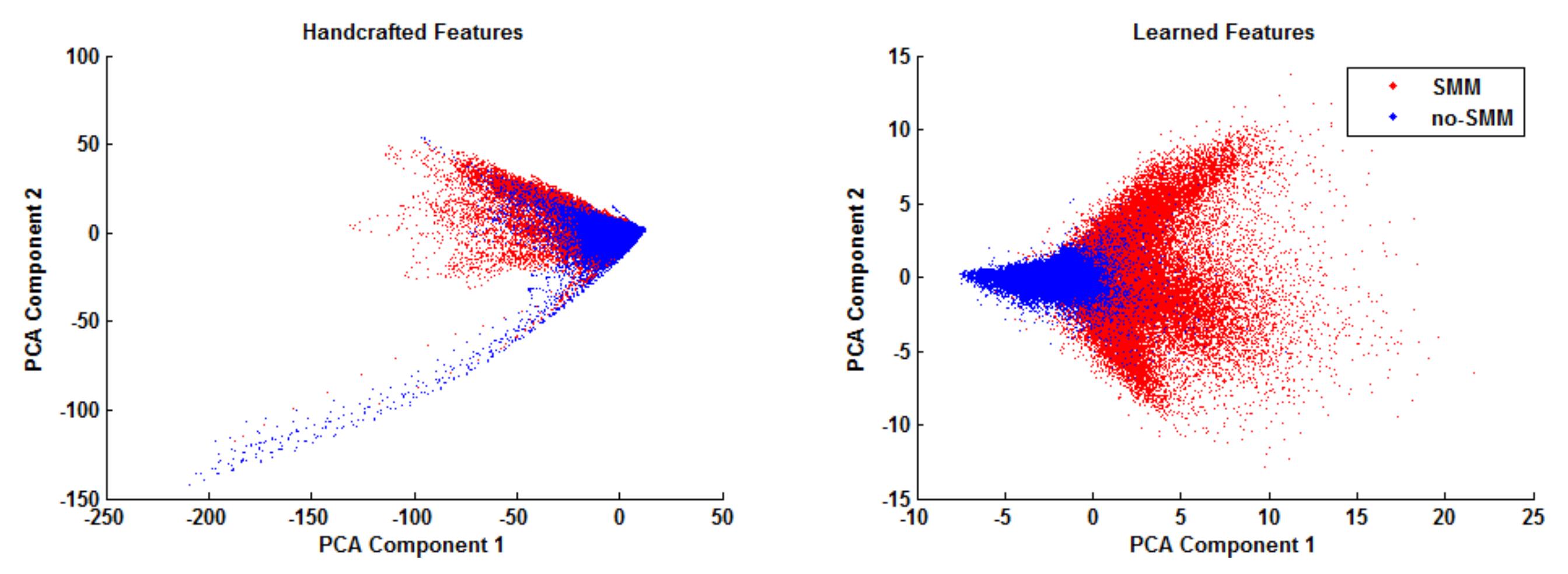}
         \end{subfigure}
         \caption{Handcrafted and learned features in 2-dimensional PCA space for SMM and no-SMM samples.} 
         \label{fig:features}
\end{figure}

Table~\ref{tab:results} and Figure~\ref{fig:results} summarize the results of four experiments. Due to the highly unbalanced sample in the test set, the evaluation is performed by computing F1-scores. The bar diagrams are representing the F1-score of four different methods on Study1 and Study2 datasets. The x-axis represents the subjects' ID (S1-S6) and the mean results over all subjects (M). In the case of the first and the second experiments, the result of experiments is deterministic in the leave-one-subject-out scenario. Therefore, no errorbar is reported. Examination of the mean performances on two datasets highlights the following remarks:

\begin{enumerate}
\item The higher classification performance achieved by handcrafted and learned features with respect to the classification on the raw data illustrates the importance of feature extraction/learning for SMM detection.
\item Comparison between results achieved by Goodwin et al. and CNN/transferred-CNN demonstrates the efficacy of feature learning over the manual feature extraction in SMM detection. To better illustrate the superiority of learned features over handcrafted features, Figure~\ref{fig:features} shows the distribution of SMM and no-SMM samples in 2-dimensional PCA space. Samples of two classes are less overlapped in the case of learned features compared to handcrafted features. 
\item Finally, our results show that transferring the learned knowledge from one dataset to another, by pre-initializing CNN, can improve the classification performance in longitudinal studies. 
\end{enumerate}

\vspace{-2em}
\vspace{-2em}

\section{Conclusions}
\label{sec:discussion}
We proposed an original application of deep learning for SMM detection in ASD subjects using accelerometer sensors. To the best of our knowledge, this is the first effort toward applying deep learning paradigm for detecting SMMs in autism. Our experimental results showed that convolutional neural network outperforms the traditional classification on the handcrafted features. This observation supports our initial hypotheses about effectiveness of embedded feature learning and transfer learning capabilities of deep framework in providing more accurate SMM detection systems. As future work, we plan to use the speed and adaptability power of the proposed framework in a real-time scenario.


\bibliographystyle{splncs03}
\bibliography{references}
\end{document}